# An AI-Enabled Framework to Defend Ingenious MDT-based Attacks on the Emerging Zero Touch Cellular Networks

Aneeqa Ijaz[1], Waseem Raza[1], Hasan Farooq[1], Marvin Manalastas[1], Ali Imran[1,2]

1 AI4Networks Research Center, Dept. of Electrical & Computer Engineering, University of Oklahoma, USA

2 James Watt School of Engineering, University of Glasgow, UK

## Abstract:

Deep automation provided by self-organizing network (SON) features and their emerging variants such as zero touch automation solutions is a key enabler for increasingly dense wireless networks and pervasive Internet of Things (IoT). To realize their objectives, most automation functionalities rely on the Minimization of Drive Test (MDT) reports. The MDT reports are used to generate inferences about network state and performance, thus dynamically change network parameters accordingly. However, the collection of MDT reports from commodity user devices, particularly low cost IoT devices, make them a vulnerable entry point to launch an adversarial attack on emerging deeply automated wireless networks. This adds a new dimension to the security threats in the IoT and cellular networks. Existing literature on IoT, SON, or zero touch automation does not address this important problem. In this paper, we investigate an impactful, first of its kind adversarial attack that can be launched by exploiting the malicious MDT reports from the compromised user equipment (UE). We highlight the detrimental repercussions of this attack on the performance of common network automation functions. We also propose a novel Malicious MDT Reports Identification framework (MRIF) as a countermeasure to detect and eliminate the malicious MDT reports using Machine Learning and verify it through a use-case. Thus, the defense mechanism can provide the resilience and robustness for zero touch automation SON engines against the adversarial MDT attacks.

## I. Introduction:

The Internet of Things (IoT) is a communication paradigm, an infrastructure upon which communication technologies, industrial domains, and smart cities are built [1]. IoT is no longer a phenomenon; it has become a prevalent system that is transforming the existing ways of communications, connections, businesses, and peoples' lives. With the emergence of its ubiquitous applicability, it is anticipated that the number of IoT devices would reach to 75.44 billion by 2025 [1]. The IoT vision is to extend the communications to anything and anywhere, with stringent transmission latency and reliability requirements. To meet the dynamic needs between the numerous and motley devices, cellular networks provide a preeminent alternative to other stacks, e.g., low power, wide area networks (LPWAN), Wi-Fi, Bluetooth etc. Cellular IoT technologies like LTE-M and NB-IoT provide deeper coverage, enhanced security and privacy features, easy remote management and analytics, hence cellular IoT networks are capable of facilitating massive flows of data around the globe. However, there is a surging demand from the heterogeneous IoT devices, the heterogeneity includes computational as well as communication capabilities. The nomadic deployment causes high Control-plane (C-plane) overhead, eventually cause distress situations and outages. The restoration of network connectivity, and dynamic network orchestration can be envisioned by self-organizing network (SON) functions. The primary objective of SON in cellular IoT networks is to optimize the network performance in terms of scalability, agility, robustness, and capacity with minimal human intervention. For automatic cellular optimization and maintenance, there is a need to continuously estimate and monitor the network key performance indicators (KPIs) like accessibility, retainability, throughput, and mobility support. For this purpose, the 3rd Generation Partnership Project (3GPP) has standardized Minimization of Drive Tests (MDT) reports from user equipment (UE) to be sent to serving BS in Release 10. The true potential of SON is contingent on the timely availability of network measurements; therefore, MDT feature is considered as a key enabler for SON functions. MDT reports consist of periodical measurements including GPS location shape: latitude, longitude, altitude, Reference Signal Received Power (RSRP), Reference Signal Received Quality (RSRQ), and Signal to Interference and Noise Ratio (SINR) values of the serving and neighboring base stations. In addition, it has information about Channel Quality Indicator (CQI), an indicator of communication channel quality. As MDT reports contain data to estimate user location through more than one method (GPS or Received signal strength-based triangulation), aggregating the MDT reported data enables the generation of the estimated coverage map [2, 3].

The augmentation of next generation mobile IoT technologies has many advantages, however, it also incurs various types of challenges and security threats. IoT devices, in the absence of additional security measures are becoming more vulnerable and easier target for adversaries. The IoT systems can be compromised by several attacks as mentioned in the literature such as eavesdropping, modification, and repudiation [4, 5]. In IoT cellular networks, a malicious user/hacker can intend to forge its identity, manipulate the transmitted MDT reports or CQI value in CSI measurements and can provide inaccurate/deceiving information to falsely trigger the SON engine, thus giving a new dimension to the security aspects in the emerging IoT cellular networks. This improvised attack, which we call *Adversarial attack*, can pose a major challenge for the stable operation of SON functions and network performance in the cellular IoT networks. Unfortunately, the existing literature considers the security of IoT systems independently, without considering the consequences of attacks on the decisions taken by the SON engine for the network optimization. To the best of our knowledge this is the first of its kind study that addresses this challenge in the area of IoT cellular networks enabled by the SON functions. This article aims to answer the three fundamental questions: *How vulnerable IoT cellular network can be? What can be the repercussions of exploitation of this vulnerability? How to manage the susceptibility of the future SON enabled cellular IoT networks against the potential adversarial attack?*

*While the discussion is framed in context of IoT cellular network, findings are equally applicable to Non-IoT cellular networks as well.*

The intent of this paper is to discuss the various types of network and software attacks in IoT systems, how these attacks have detrimental impact on different network segments, i.e., end users, access, or core network. Particularly, how an adversary can exploit the MDT/CQI reports to launch the adversarial attacks. The main contributions of this paper can be summarized as:

1) We introduce an impactful adversarial MDT attack and analyze the cascade of its detrimental impact on the cellular network performance.
2) We address the repercussions of the adversarial attacks on the SON functions and demonstrate how the fake MDT/CQI reports can create conflicts among various SON functions, undermining the self-optimization capability of the system.
3) To detect the adversarial attack, we propose a novel multi-module Malicious MDT Reports Identification framework (MRIF) as a countermeasure that aims to detect and distinguish false MDT reports from the true reports based on machine learning (ML) techniques.
4) We demonstrate the effectiveness and robustness of the proposed defense mechanism against the MDT attack using a detailed use-case for cell outage compensation (COC) SON function and experimental results.
5) As an optional layer of verification, we touch upon a new concept of Flying Drive Testers (FDT) with on-board UEs to increase the resilience of zero touch automation enabled by the SON engine.
6) Finally, some open challenges are discussed for the implementation of ML-based security measures in the IoT systems to trigger the research efforts in this emerging wireless communication networks.

## II.     Attacks in IoT:

The security and privacy preservation of IoT cellular networks is a formidable challenge due to the diversity, complexity, and enormous interconnected resources. There are numerous sophisticated attacks that an adversary can perform on IoT network such as eavesdropping, modification, tampering, and repudiation, due to the lack of efficient tracking mechanisms, this false information may lead to users' inconvenience and serious consequences.

There are several attacks encountered by the IoT devices, based on the vulnerability, occurrence, damage level, and detection chances in various network segments (UE, access network, and core network), as shown in Fig. 1. Based on the vulnerabilities in the sensing, network, and application layer in IoT networks, the attack can be classified into four categories: 1) *physical attacks, 2) network attacks, 3) software attacks, and 4) encryption attacks* [5]. In *physical attacks*, the adversary targets the hardware of the devices. The adversary to identify new vulnerabilities, access the device by buying a copy of the IoT device, performs the reverse engineering and creates a false attack to observe/analyze the estimated outcomes. In *network attacks*, the malicious node tries to compromise the connectivity between the device or the connection between device and cellular network. Meanwhile, in *software attacks* the adversary focuses on the denial of services and data stealing by employing the spyware, viruses, or worm. Lastly, in *encryption attacks,* the focus is to destroy the encryption technique, to obtain the private key of the network. As Fig. 1 illustrates that various IoT attacks can target different network segments, i.e., the target of some network and software attacks are the end user devices, whereas other network and software attacks are specifically designed for the access and core networks. Moreover, there are some attacks that can be launched simultaneously in more than one network segment, making them difficult to track, compared to the other attacks. For instance, backdoor attacks, DoS attacks, and malicious script attacks can co-exist in multiple network segments, making it harder to identify the attack and launch the defense mechanism accordingly. Numerous countermeasures are proposed in the literature for the network and software attacks [6].

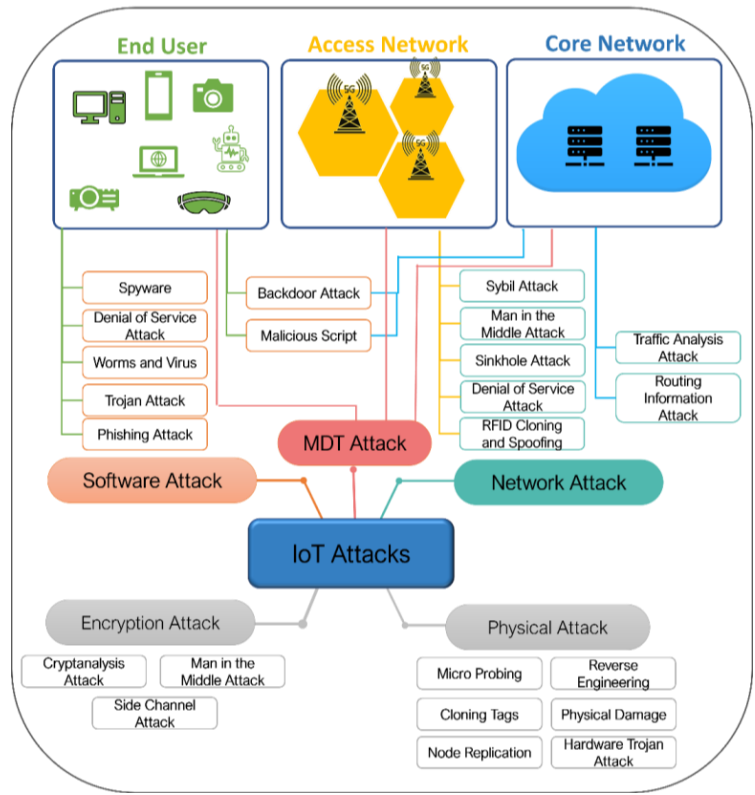

Figure 1: Taxonomy of the IoT attacks

In the context of cellular IoT networks, one form of attack we introduce in this work is "MDT attack". In contrast to other types of IoT attacks having an impact on either one or two network segments, this newly introduced MDT attack has its deleterious impacts on several network segments. For instance, when the adversary takes charge on the registered IoT device and start sending malicious MDT/CSI reports to send false data to the base station (BS), hence attacking the access network. When the SON engine performs the network optimization based on the misleading MDT/CSI reports, the impact of this attack may propagate to wider radio access network (RAN) as well as the core network. Hence, the disseminating nature of this attack in all the three network segments, makes it distinctive and more virulent compared to the other network and software attacks. In addition, with more and more focus in transitioning open RAN to the cloud, the threat of this attack will increase many folds. Therefore, robust countermeasures are required for secure and resilient IoT cellular networks.

### III. MDT reports and adversarial attack using malicious MDT reports:

MDT reports are vital for network monitoring and optimization, as the IoT and other UE devices periodically update the network about the signal condition, providing information of spatial and temporal data related to the QoE in a specific area. Through MDT measurement configurations, network can instruct the UEs to log network measurements, such as RSRP, RSRQ, and SINR of the serving and neighboring cells, and send them to the serving BS through Radio Resource Control (RRC) signaling messages. Thus, MDT eliminates the need of physical drive test for the system quality assessment. Furthermore, compared to drive test-based performance assessment that can gauge performance only on paved roads, MDT reports can not only provide more spatially holistic picture of the network performance but also they offer opportunity for more timely coverage analysis, user experience, and anomaly detection in the network performance. The MDT scheme supports two reporting configurations: (i) *immediate reporting* wherein the UE instantly reports the measured radio conditions when the preconfigured triggers are met, and (ii) *logged mode* wherein the UE stores measurements and reports them upon the expiration of a periodic timer. MDT reports are tagged with location information of the reporting UEs. These attributes play a significant role for facilitating the SON algorithms to take necessary actions for the optimal network performance [7].

### a. How vulnerable is IoT cellular network to MDT-based attacks?

The true potential of SON is contingent to the timely and accurate availability of network MDT measurements. Based on the intelligence extracted from the reports, SON engine initiates appropriate SON-functions to achieve optimum network performance that aligns with the business tailored objectives prescribed by the operator. Effectively, the MDT feature is a pre-requisite for enabling a range of diverse 3GPP standardized SON functions such as mobility load balancing (MLB), cell outage compensation (COC), coverage and capacity optimization (CCO), enhanced inter-cell interference coordination (eICIC) and Energy Efficiency (EE) to name a few [8].

However, there lies one caveat here - if we look from the security perspective, SON engine optimizes the network by relying on the information sent by the UE which is in direct reach to the public. This inherently can serve as a backdoor for the hackers to influence the SON mechanism by sending modified/false MDT measurements within *UEInformationResponse* RRC messages back to the network through the compromised/hijacked UEs. Even the sophisticated UE devices such as smartphones, which have additional security softwares, have vulnerability due to the availability of open source baseband stack like OsmoComBB [9], that can be run on compatible phones. OsmoComBB allows obtaining of general understanding of the structure of cellular stack, which can be very helpful for binary analysis and reverse engineering of baseband stacks, commonly available in the market. Compared to smartphones based UEs, however, it is the inexpensive IoT devices having limited intelligence and no additional security layer, which makes them more vulnerable to the MDT-based attacks, thereby providing core motivation for this study. For conciseness, here forth, we will use the term UE to represent all type of cellular and other devices including IoT.

Hackers can exploit these vulnerabilities of UE to improvise cellular network signaling attack. More specifically, in the context of MDT adversarial attack, hackers can modify the baseband stack to embed deceiving measurements in the *UEInformationResponse* RRC messages and send them back to the SON engine of the network. If the malicious UE succeeds in modifying protocol stack of the baseband processor, then a successful attack can be executed, as state-of-the-art SON engines are designed to trust whatever data it receives from that UE in the network.

### b. What can be the repercussions of MDT/CSI-based attacks?

If a network attacker succeeds in modifying the network stack in a UE then they will be capable of sending customized false MDT reports back to the SON engine. These deceiving reports can force SON engine to erroneously trigger several SON functions that can have devastating impact on the entire network performance. For instance, hacked UEs can falsely report MDT reports with very low RSRP values from areas which actually have good coverage, thus can trigger COC SON function in the network when and where it is not needed. This may lead to an unstable network condition forcing the network to drift into a sub-optimal configuration state, as shown in Fig. 2. The figure shows that in case of actual outage detection Fig 2 a), the COC SON function is triggered and increases the transmit power of neighboring cells to compensate the UE previously connected to the cell in outage as shown in Fig 2 b), hence the system SINR is restored to quite an extent, as demonstrated in Fig 2 c). However, if outage compensation mechanism is activated based on malicious MDT reports, (Fig. 2 d and e). The SON function is triggered erroneously, causing a decrease in the network SINR because of the increase in the interference caused by the unnecessary increase of transmit power of the compensating BS. This negative affect of MDT-based attack that leads to erroneous triggering of COC by SON engine can be seen in Fig 2 f). This use-case exhibits how an attacker by sending the false MDT report to a SON engine, can mislead SON engine to erroneously trigger/halt a SON use-case and thus jeopardize the network performance.

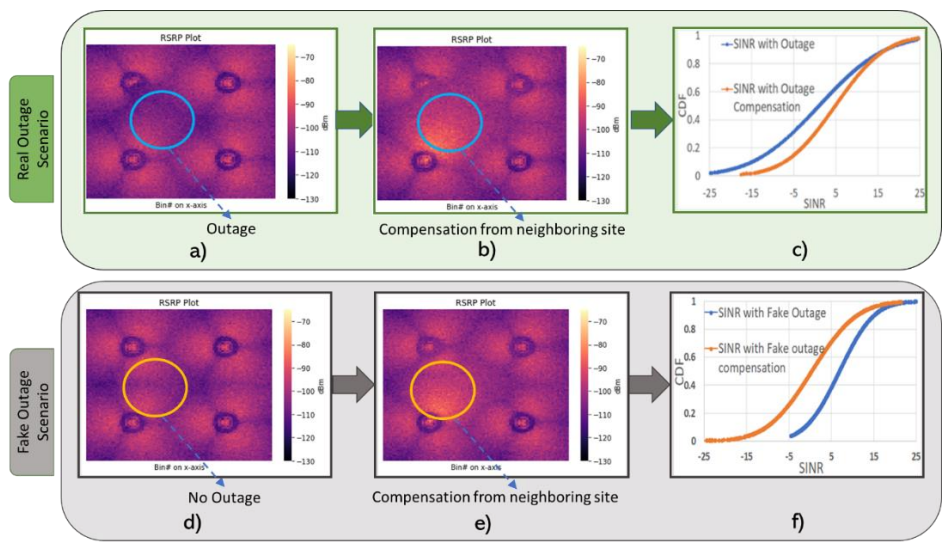

Figure 2: Adversarial MDT attack impact on the system's SINR and network performance

An adversarial attack of this kind can target any SON function as shown in Table I. For instance, a compromised UE can send malicious reports related to handover parameters that can erroneously commence Mobility Robustness Optimization (MRO) SON function. Same is true for other SON functions such as MLB and EE. In fact, a specially crafted coordinated attack by the compromised UEs can initiate any desired SON function in the network. Table I shows the potential impact of the adversarial attack on the 3GPP defined SON functions and identifies the associated negative impact on the network KPIs. We categorize the SON functions which are most likely to be negatively affected from the MDT attacks. In case of *centralized SON*, the local impact of MDT attacks can cause interactions at central SON server that can have an influence on the global behavior of the network. In case of *distributed SON*, in addition to local performance degradation, the unnecessary call for SON functions caused by MDT attacks can lead to the wastage of network resources like computational power, signaling bandwidth, etc. Moreover, it has been highlighted in the literature that multiple uncoordinated SON functions are subject to a large number of potential logical, parametric, or objective conflicts [10]. Therefore, even with distributed SON, local MDT attack can lead to the sub-optimal performance at a larger scale. Furthermore, coordinated attacks by compromised UEs can be configured such that the network never reaches to its optimum stable point, rather it keeps on oscillating between different states having negative impact on the operational expenses

(OPEX) and QoS. Therefore, it is crucial to devise a robust defense mechanism against these MDT-based adversarial attacks.

TABLE I: Implication of MDT attacks on SON Functions

| SON Function | Potential Attack Description | Affected KPIs | Severity of Effect when Attacked |
|---|---|---|---|
| MLB | Adversarial UEs can report poor SINR and request for higher throughput. To accommodate the request, BS will allocate more Physical Resource Blocks (PRBs) which might trigger false MLB | • PRB Utilization<br>• Traffic Volume<br>• MIMO Rate<br>• Carrier Aggregation Activation<br>• Accessibility<br>• Handover Success Rate<br>• Downlink Throughput | High |
| MRO | Handovers mainly depend on the RSRP measurement of the source and target cell. An attacker can trigger too early or late handovers, leading to increased handover failures. As a compensation, MRO might add unreasonable CIOs | • Handover Success Rate<br>• Downlink Throughput<br>• Retainability<br>• Integrity | High |
| COC | Fake outage can cause serious issues in the network if undetected. The BS might activate COC to increase the power or change the tilt of the BSs to cope with outage, which affect the SINR of the BS under attack | • Accessibility<br>• Retainability<br>• Handover Success Rate | Medium |
| CCO | The adversarial UE can report incorrect measurement, e.g. very poor RSRP or RSRQ. In return, BS increase its coverage which might increase interference and overshooting with other neighboring cells | • Coverage<br>• Capacity<br>• Handover Success Rate | Medium |
| eICIC | The malicious reported RSRP or RSRQ can falsely trigger the eICIC function. Based on that the BS can unnecessarily change the coordination of physical resources between neighboring cells to reduce interference | • Capacity<br>• Integrity<br>• PRB Utilization | Medium |
| EE | Some BS are configured to turn off during certain period (non busy hours) to conserve energy. An attacker might fool these BSs and not letting it sleep, thinking some UEs need connection by sending false request or measurements | • Energy Efficiency<br>• Availability | Low |

**MLB:** Mobility Load Balancing, **MRO:** Mobility Robustness Optimization, **COC:** Cell Outage Compensation, **CCO:** Coverage and Capacity Optimization, **eICIC:** enhanced Inter-Cell Interference Coordination, **EE:** Energy Efficiency.

## IV. A framework to counter adversarial MDT Attacks

Adversarial attacks have the tendency to take control over the IoT devices and propagate the impact of the attack into different networks segments by sending false MDT reports, leading to sub-optimal performance. Fake MDT reports can be logged either from the registered and unregistered end-user devices which makes the detection challenging. This inherent vulnerability of the network demands the need of devising countermeasures against such pernicious attacks. In this section, we propose a defense mechanism called Malicious MDT Reports Identification Framework (MRIF) to detect the deceitful intruders, as illustrated in Fig. 3. The breadth and depth of MRIF anchor on its capability to detect outages and distinguish fake MDT/CSI reports from the real outage reports. The framework consists of the following modules:

*a)* **Anomaly Detection Module (ADM):**

This module receives the MDT reports from various users and detects any anomalies in the reported MDT/CSI data. One example of such anomaly which attackers can exploit is the cell outage. ADM leverages ML-based anomaly detection algorithms such as Cell Outage/Degradation Detection SON algorithms [11]. Anomaly detection algorithms are commonly based on classification, nearest neighbor, clustering, and statistical measurements, to detect and identify the anomalous MDT reports that imply a degraded performance due to some possible fault in a cell. Some of the techniques traditionally used for anomaly detection are One-Class SVM and LOF [12, 13]. Recent studies [13, 14] also demonstrate that leveraging XGBoost, Autoencoder gives better outage detection and more resilience to various scenarios.

For our proposed case study, we have considered the XGBoost, and AE as discussed in Section V. Leveraging these models enable the separation of normal MDT reports from the outage data, with satisfactory accuracy.

### b) Malicious Reports Filtering Module (MRFM):

To further filter out the malicious MDT reports indicating the false outage from the genuine MDT/CSI reports indicating true outages, the outage data is fed into the MRFM. The working of this module is premised on the notion that the malicious reports from a compromised UE will differ from the genuine anomalous MDT reports which are in the same vicinity. This observation stems from the fact that the shadowing phenomenon has a significant impact on the reported MDT results, particularly in an environment where it remains correlated over small distances. Therefore, if the number of malicious reports in the neighboring (MRN) region is greater than a threshold it indicates actual outage which might be due to technical or operational discrepancy in the cell or network. A real outage cannot be an isolated event for single or few users in a particular region, rather a significant number of users from the same region should report it as the outage event. In the LOF scheme [13], one hyperparameter, i.e., "n_neighbor" $\eta$ (threshold) is essentially capturing the essence of this observation. Hence, after performing the hyperparameter tunning, we are able to get the good performance for $\eta = 10$, i.e., if less than 10 neighbor UEs report poor MDT in a particular region then it manifests an adversarial attack, which depending on the severity will either require an additional verification process or immediate blocking of the compromised adversary sending false MDT/CSI reports.

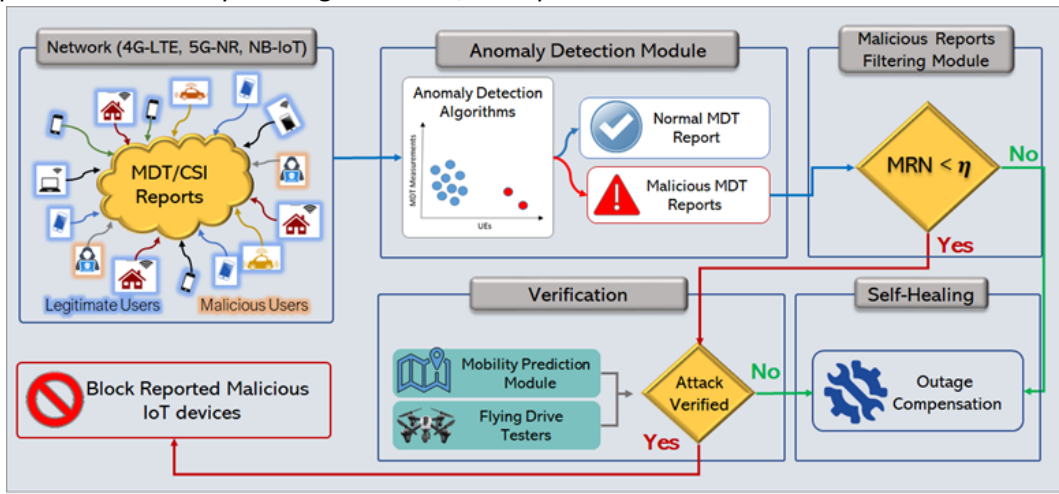

Figure 3. Malicious MDT Reports Identification Framework

### c) Verification of adversarial attacks using Flying Drive Testers:

In addition to this multi-module MRIF defense mechanism, to provide an additional layer of protection and make the framework more rigorous/robust, we suggest that Flying Drive Testers (FDTs) can be leveraged by the network operators. When the sparse fraction of the anomalous reports is reported from a cell, then the network can dispatch the nearest FDTs to the reported location to verify the MDT readings through its onboard UE. Aerial movement allows the possibility of testing at any location in minimum amount of time with reduced OPEX, compared to drive testing that can be limited to only paved locations. These FDTs can be kept at different cell sites distributed across the network. An optimization problem can be formulated and solved to identify the optimum locations and number of FDTs in each site such that the time to reach any location is minimum. When the sparse fraction of anomalous reports is reported from a cell, then the network can dispatch the nearest FDTs to the reported location to verify the MDT readings through its onboard UE. Based on the readings from the FDTs, network can decide whether to initiate self-healing algorithms or to flag the UEs reporting anomalous reports as compromised UEs and block their MDT reports from reaching SON engine. However, this extra verification time through FDTs comes with associated costs, i.e., it might affect the user's QoE in case of actual outage.

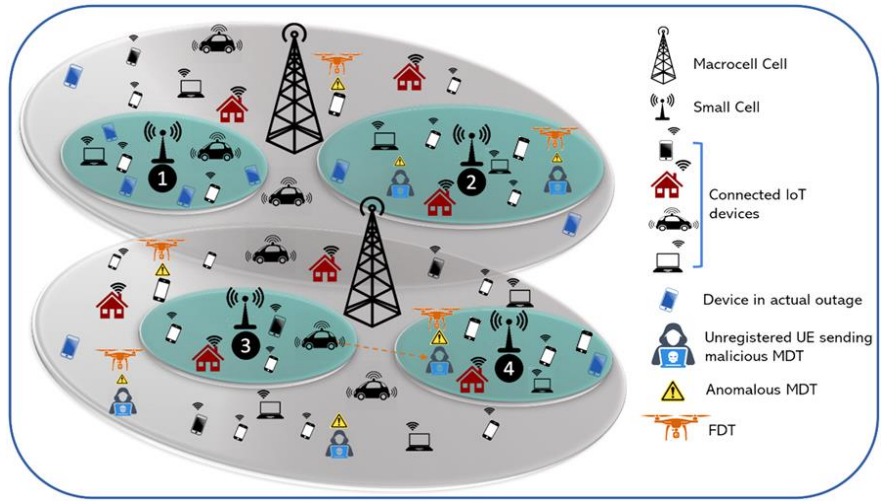

Figure 4: Network model with the deployed Flying Drive Testers (FDT) for anomaly detection. Small cell 1 has many IoT devices reporting the actual outage through anomalous MDT reports. While malicious MDTs identified in small cells 2 and 4. As in cell 2, very small fraction of MDTs is anomalous therefore they are identified as attackers. In case of cell 4, network is performing verification through IoT devices from small cell 3 and the FDT.

## V.     A Case Study to show the efficacy of the proposed MRIF:

We demonstrate the working of the proposed multi-module defense mechanism with a use-case of identifying the adversarial MDT attack in the presence of real outage MDT reports. We aim to demonstrate that the ML-enabled framework can identify fake MDT data sent by the adversary IoT devices from the genuine outage reports. The subsequent discussion and Fig. 5 show the efficacy of the proposed MRIF in detecting the malicious MDT-based attack:

- We utilize a 3GPP compliant system-level simulator SyntheticNet [15] to model the network and generate the MDT data for the outage detection. The relevant parameters are also set according to the standards, given in Fig 5 a).
- The outage scenario is simulated by setting the transmit power of three BSs to zero. The resulting network coverage map with 3 BSs in outage is shown in Fig 5 b). In the area of 1 $km^2$, 1000 users are deployed where the UEs associated to these 3 BSs report real outage by sending low RSRP values in their MDT reports. Whereas, to simulate the adversarial attack we reside to the idea that any user can act as a potential adversary and falsely reports poor signal condition, despite being connected to the normal BS. Thus, the MDT data is comprised of two main categories, i.e., normal MDT reports and outage MDT report. The category of outage reports further constitutes of i) real outage, and ii) malicious

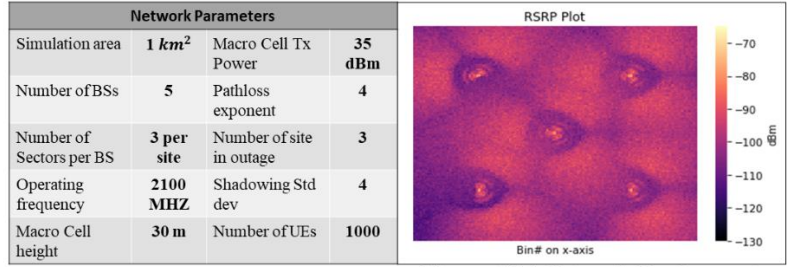

Figure 5 (a) Simulation Parameters

Figure 5 (b) Network Deployment using System Level Simulator

|  | Outage Severity | | | | | |
|---|---|---|---|---|---|---|
| **MDT Data Size** | 01-Cell-Outage | | 02-Cell-Outage | | 03-Cell-Outage | |
|  | AEN | XGB | AEN | XGB | AEN | XGB |
| Low=2500 | **0.952** | **0.921** | 0.932 | 0.909 | 0.914 | 0.906 |
| Med=5000 | 0.953 | 0.973 | **0.950** | **0.939** | 0.934 | 0.932 |
| High=7500 | 0.987 | 0.969 | 0.962 | 0.951 | **0.971** | **0.966** |

Figure 5 (c) Performance comparison of the ADM as the F1-score for different cases of MDT Data Size and Outage Severity

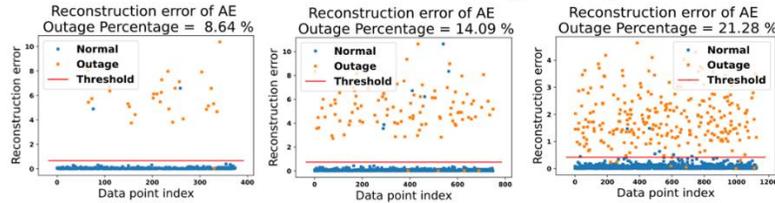

Figure 5 (d) Reconstruction error plot for the selected combination of MDT data density, and outage severity.

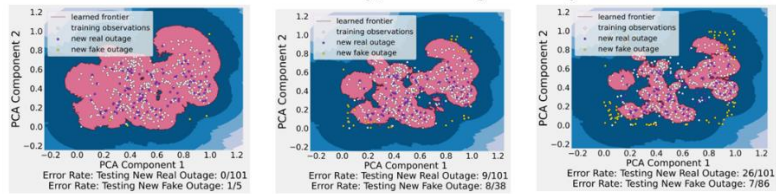

Figure 5 (e) Comparison of real and fake outage separation for MRFM (varying degrees of fake outages).

Figure 5. Use-case showing the efficacy of the MRIF framework.

outage MDT reports. The MDT measurement report constitute of the features, i.e., longitude and latitude of the users, RSRP, and RSRQ of the serving base station. To incorporate the correlation information from the neighboring cells we have also included the RSRP and RSRQ from the six strongest neighboring BSs as features. This data is fed into the ADM with the goal of developing an optimal model that can distinguish the normal reports from the malicious MDT reports. Among the aforementioned ML-based algorithms, we implement the AE and XGBoost anomaly detection due to the robustness, and sensitivity to the higher shadowing data. The autoencoder is trained on the normal MDT data and tested using the composite data containing normal and outage MDT reports. The regenerative capability of the autoencoder is measured by the regeneration error, separated by selecting the appropriate reconstruction error threshold (red horizontal line) as shown in Fig. 5 (c). For AE training and testing, we generate labeled synthetic MDT data using SyntheticNet [15] (For real implementation of the proposed solution either a simulator or a digit twin of the network combined with the real

MDT reports can be used). Using labeled data make the estimation of anomalous outcomes and score comparable to that of the supervised classification. To effectively detect these anomalous outcomes, one of the key parameters of AE is the reconstruction error threshold. This error is expected to be marginal for the normal instances and exceptionally higher for the outage instances. Lowering this threshold may result in the detection of more anomalies, but it may also cause normal instances to be misinterpreted for malicious ones i.e., higher false positive rate. On the other hand, raising this threshold prevents false positive detection at the expense of detecting lesser anomalies. Several approaches to determine the reconstruction error threshold exist with varying complexity. These methods include manual selection based on the reconstruction error distribution, utilizing mean and standard deviation of the reconstruction error, and leveraging the $q^{th}$ quantile of the reconstruction error. For our anomaly detection framework, we employ the $q^{th}$ quantile of the distribution of reconstruction error, where the parameter $q$ is determined by the ratio, i.e., the number of outage instances relative to the total number of instances in the training or test datasets. This method of tying the threshold to the $q^{th}$ quantile of reconstruction error and making q adaptable to the number of outage occurrences enables the framework to function effectively in scenarios with varying outage severity. For module 1 experiments, we have considered different combinations of MDT (normal and outage) dataset size. We considered the effectiveness of the outage detection model with the increase in the overall data size, and the percentage of outage instances in that data while the percentage of fake outages remains constant, i.e., 1% of the MDT reports size. More specifically, here we consider the low, medium, and high density of MDT data size. We have also considered different cell outage scenarios, i.e., 1, 2, and 3 cells-based outage severity out of 9 cells. To compare the performance of the outage detection module, AE is evaluated against the XGBoost classification method. We get the results for the AE and XGBoost models for the F1-score as shown in Fig. 5(c) and reconstruction errors in Fig. 5(d), respectively. It can be observed that for the same data size, there is a degradation in the AE and XGBoost performance, with an increase in outage severity. This is because, given the increase in the outage percentage, the chances of misclassification increase, specifically for the AE scheme which is designed to separate the anomalies. Hence the performance degradation is relatively less than the XGBoost scheme. It is worthnoting that despite the performance differences discussed above, overall, both schemes in ADM perform reasonably well for the anomaly detection, i.e., to separate the normal and outage instances.

The next step is to separate the fake outages from the real outages by exploiting MRFM. We exploit user's geographical locations, mapped with the corresponding MDT reports as inputs for training the local outlier factor (LOF) model [13]. The rationale behind the use of LOF and geographical positions stem on the fact that user reported real outages are usually geographically collocated and associated with the same BS. The LOF utilizes the density of the points in the neighborhood for the outlier detection, and it can be trained to exploit the geographical co-location to separate the real and fake outages as shown in Fig. 5 (e). The LOF model is trained on the real outages data, with parameter settings of 15 neighbors, 0.15 contamination and Euclidean distance, to learn the frontier as shown in the figure. It can be observed that with these settings, LOF is able to correctly predict 86% of the real outage instances and 88% fake outage instances. For the evaluation of second module MRFM of the proposed framework that is designed to detect fake outages from the real outage MDT/CSI reports, we also defined three different cases of real and fake outage percentages and test the LOF scheme's efficacy. We test the model performance when the fake outage percentage takes values from low to medium and high cases, and the real outage percentage remains constant. For this module, we consider 2-cell outage scenario, however, instead of a fixed fake outage percentage, here we changed the malicious outage rate from 5% to 90%. The results for this case are listed in Fig. 5 (e).

The error rate, given as the ratio of the number of misclassified labels to the total number of labels in a particular class is given on the x-axis. Here, along with the MDT data, the geographical location of the MDT-generating UEs is exploited to separate the real outages from the malicious/fake outages. It is observed from these results that the PCA-assisted LOF-based separation in the MRFM learns the frontier of the real outages and is able to correctly filter the large percentage of fake outages based on the learned frontier.

- The results verify that the proposed MRIF defense mechanism can detect adversarial attack by leveraging ML for outage detection and further identification of the malicious outages with high accuracy. Eventually, enabling the network to block and restrict these attackers from causing any sub-optimality in the network.

## VI: Open Research Challenges:

In this section, we discuss the open research challenges faced due to the unavailability of representative IoT cellular networks data and in the implementation of ML techniques:

- *Data sparsity*: For ML security model training, we rely on the anomalous data which is very rare to find. Thus, it is likely to train the ML model on a highly imbalance data as the attacks are sparse events. Training the model using such imbalance datasets can significantly impact the performance of the classifiers. Sparsity impacts the training of ML models, however, for this work, we have not considered the sparsity problem and assume that the synthetic data generated using a system level simulator is enough to train the model. However, sparsity challenge has been well studied in the literature. To address the sparse data challenge classic interpolation techniques such as matrix factorization spline, Generative Adversarial Networks (GANs), transfer learning, and deep learning techniques can be implemented.

- *Concept drift:* It represents the fact that even if we succeed to observe all the features of the network, for the training of the ML model, for its normal behavior to classify the anomalous behavior (outage, performance degradation), the system is time variant, and it keeps on evolving. Due to the non-stationary nature of the network features they keep changing over the time, and such trained model can be easily tampered by the attackers. To overcome this challenge the network should be trained on the large scale of spatiotemporal data.

## Conclusions:

In this article, we have presented first of its kind minimization of test drive (MDT)-based adversarial attack that have a cascade of detrimental impact on the network performance of the cellular network in general and IoT cellular network in particular. The attack can be launched by compromising the MDT reports, that are poised to form the basis of AI-based network automation. Thus, the compromised users can trigger the unnecessary activation/deactivation of SON or other zero touch automation functions. We have demonstrated that this unwarranted activation of SON function can cause severe network performance degradation. To combat such deleterious MDT-based attack, we present a novel and effective Malicious MDT Reports Identification framework (MRIF), that can detect and eliminate the malicious MDT reports using Machine Learning (ML). The use-case of cell outage compensation SON function demonstrates that the defense framework can detect and distinguish false MDT reports from the true reports-based ML techniques with high accuracy. Thus, using a two-tier anomaly detection and malicious reports filtration,

MRIF provides a promising security for the emerging cellular networks, empowered by SON functions that relies on MDT data. In conclusion, this framework is capable to provide an efficient and rigorous solution against MDT-based adversarial attacks, as an interim coverage compensation for the users affected by a real network outage in the next generation cellular networks.

**Aneeqa Ijaz**

Aneeqa Ijaz holds a B.Sc. degree Electrical Engineering from the University of Engineering and Technology, Lahore, Pakistan, M.Sc. degree in Telecommunications and Computer Networking from NUST, Pakistan. She is a Fulbright scholarship recipient for her Ph.D. in Electrical and Computer Engineering at the University of Oklahoma, USA. Her research interests include AI-based solutions for wireless communication and healthcare domain.

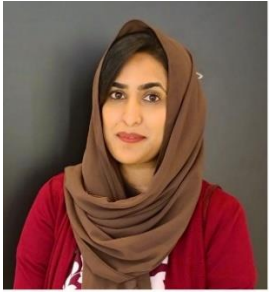

**Waseem Raza**

Waseem Raza received the B.Sc. and M.Sc. degrees in Telecommunication Engineering from the University of Engineering and Technology at Taxila, Pakistan, in 2014 and 2016, respectively. He is pursuing his Ph.D. in Electrical and Computer Engineering from the University of Oklahoma, USA. His research interests include AI and ML techniques for 5G and beyond wireless networks.

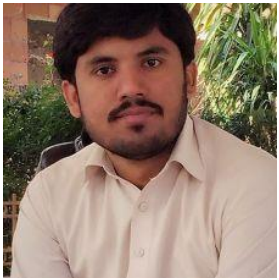

**Hasan Farooq**

Hasan Farooq is Senior AI Researcher at Ericsson Research, USA. His background is AI/ML driven zero-touch automation algorithms for Radio Access Networks. He holds a Ph.D. degree in Electrical and Computer Engineering and Post-Doc from the University of Oklahoma, USA. He has authored/co-authored over 50 publications in high impact journals, book chapters, and proceedings of IEEE flagship conferences on communications. He also has patents in the area of SON algorithms.

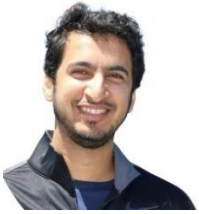

**Marvin Manalastas**

Marvin Manalastas, holds a B.S. degree in electronics engineering from the Polytechnic University of the Philippines and an M.S. degree in electrical engineering from The University of Oklahoma (OU). He is currently pursuing a Ph.D. degree in electrical engineering at OU with research focus on applying machine learning to optimize 5G and future networks.

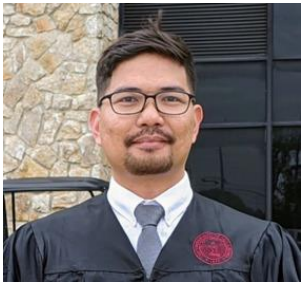

**Ali Imran**

Ali Imran (Senior Member, IEEE) is Professor of Cyber Physical Systems in James Watt School of Engineering, University of Glasgow. He is currently on leave from University of Oklahoma where he is Williams Presidential Professor in ECE and the founding director of the Artificial Intelligence (AI) for Networks (AI4Networks) Research Center. His research interests include AI and its applications in wireless networks and healthcare. His work on these topics has resulted in several patents and over 150 peer reviewed publications. He is an Associate Fellow of the Higher Education Academy, U.K.

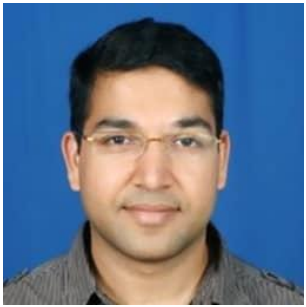